\documentclass[letterpaper, 10 pt, conference]{ieeeconf}  

\IEEEoverridecommandlockouts                              

\usepackage{graphics} 
\usepackage{epsfig} 
\usepackage{mathptmx} 
\usepackage{times} 
\usepackage{amsmath} 
\usepackage{amssymb}  
\usepackage{xcolor}
\title{\LARGE \bf
Where Will They Go? Modelling Multimodal Pedestrian Manoeuvres from Ego-centric Videos
}

\newcommand{\rev}[1]{{#1}}

\author{Yuxuan Xie$^{1}$, Nicolas Pugeault$^{1}$, Chongfeng Wei$^{2}$, Hubert P. H. Shum$^{3}$, and Edmond S. L. Ho$^{1*}$
\thanks{*Corresponding author, email: {\tt\small Shu-Lim.Ho@glasgow.ac.uk}}
\thanks{$^{1}$School of Computing Science, University of Glasgow, Glasgow, United Kingdom}
\thanks{$^{2}$James Watt School of Engineering, University of Glasgow,  United Kingdom}
\thanks{$^{3}$Department of Computer Science, Durham University, Durham, United Kingdom}
}

\begin{document}

\maketitle
\thispagestyle{empty}
\pagestyle{empty}

\begin{abstract}
Pedestrian trajectory prediction from an \rev{on-board} ego-centric camera is challenging since it depends on complex interactions with vehicles and scene context, as well as the intention of the pedestrian. The task becomes even more challenging  since pedestrian intention is often ambiguous from historical observations alone, leading to an inherently multimodal distribution over future trajectories. Existing CVAE-based predictors learn a latent representation by aligning the prior conditioned on trajectory history with the posterior conditioned on both history and future trajectories. This alignment may suppress multimodality in the latent space, leading to mixed-mode predictions that interpolate between distinct future behaviours. In this paper, we propose MMPM, a mode-aware framework that separately models future trajectory distributions into semantically meaningful modes based on the pedestrian's crossing behavior. MMPM consists of two modules: behavior-aware Pedestrian Interaction Module (PIM) that jointly captures pedestrian–vehicle and pedestrian–environment interactions by introducing gaze and hand gesture, and a CVAE-based Mode-aware Trajectory Predictor (MTP) module to model the future trajectory distributions in two modes, crossing and non-crossing the road, separately. A query-based decoder further enforces mode consistency during decoding. Experiments on PIE and JAAD datasets show that our method surpasses state-of-the-art baselines. Our proposed MTP is model-agnostic, which can be integrated into existing frameworks such as BiTrap and SGNet to further improve future trajectory prediction performance. We additionally introduce a data-driven validation protocol that retrieves spatio-temporally consistent real-world trajectories from the same scene and evaluates predictions against these retrieved examples, demonstrating improved frame-wise displacement errors over previous work.
\end{abstract}

\section{INTRODUCTION}

Pedestrian trajectory prediction is one of the most important tasks in autonomous driving. 
In recent years, this problem 
has become an active research area and 
has achieved considerable progress by learning key factors that influence pedestrian motion, including interactions among agents~\cite{Rasouli2020BifoldAS}, the impact of surrounding context~\cite{9636241,zhai2022social}, and goal-driven behaviors~\cite{Yao2020BiTraPBP,Wang2021StepwiseGN}.  

To model the dynamics of the interactions~\cite{Crosato:Survey} between the pedestrian, other road users and the environment, the historical features of agents are typically aligned and processed using attention~\cite{10569027,Su2022CrossmodalTB,Yuan2021AgentFormerAT,Saadatnejad2023SocialTransmotionPH} or recurrent models~\cite{Salzmann2020TrajectronDT,Ivanovic2018ModelingMD} to capture the spatio-temporal information. Visual inputs, e.g. scene-level images~\cite{Neumann2021PedestrianAE} or semantic segmentation~\cite{Rasouli2022PedFormerPB,Rasouli2020BifoldAS}, are commonly used as
spatial clues for spatio-temporal modeling. Goal-driven approaches~\cite{Chiara2022GoaldrivenSR,Yao2020BiTraPBP,Wang2021StepwiseGN} 
guide the decoding process by predicting pedestrians' final goals. However, pedestrian intention is often ambiguous from historical observations alone, leading to an inherently multimodal distribution over future trajectories. 
On the other hand, encouraging results have been reported in modeling multimodality for vehicle trajectory prediction~\cite{Choi2022HierarchicalLS,wang2025c2f}.

To the best of our knowledge, none of the aforementioned approaches explicitly model the 
multimodality of pedestrian trajectories in ego-centric datasets: \textit{Given the same or similar historical trajectories, a pedestrian may have multiple possible future trajectories corresponding to distinct goals, with distinct motion patterns.} Existing stochastic prediction methods~\cite{Wang2021StepwiseGN,Yao2020BiTraPBP,Gu2022StochasticTP} employ a conditional variational autoencoder (CVAE) framework that regularizes a history-conditioned prior towards a posterior conditioned on both the trajectory history and future observations. Although the KL divergence is computed for each training sample, the shared prior is implicitly required to match the aggregate posterior induced by all valid futures corresponding to the same trajectory history. Since this aggregate posterior is inherently multimodal, representing both distributions with a single Gaussian introduces an aggregate posterior mismatch, which may encourage latent mode mixing and consequently produce averaged or less diverse future trajectory predictions.



In this paper, we propose a model, namely Modelling Multimodal Pedestrian Manoeuvres (MMPM), that models the distribution of the future pedestrian trajectories in each \textit{mode} individually. \rev{We treat the pedestrian action, i.e. \textit{crossing} or \textit{non-crossing} as the modes in this study, as this pair is binary and \textit{crossing} trajectories and \textit{non-crossing} trajectories have distinct future maneuvers.} Our framework consists of two interacting components: a Pedestrian Interaction Module (PIM) and a CVAE-based Mode-aware Trajectory Predictor (MTP). Specifically, PIM aims to learn a representation based on the interactions among the pedestrian, other road users, the autonomous vehicle and the environment. We also enhance the representation by incorporating the gesture and gaze of the pedestrian as behavioral features, which have not been explored in prior approaches, to provide additional cues for our model to predict the pedestrian's intention. Next, during the training stage of MTP, we use the ground-truth actions as the modes and we model the distribution of future trajectories under each mode separately. At inference, MTP starts by predicting the \textbf{mode} based on the output of PIM, then samples future trajectories from the learned prior for each mode according to the ratio of predicted crossing probability.


Experimental results show that the proposed method surpasses the state-of-the-art methods~\cite{Chiu2025TemporalAW,Rasouli2023ANB} on both the benchmark PIE and JAAD datasets. Our proposed MTP is model-agnostic, which can be integrated into existing frameworks such as BiTrap~\cite{Yao2020BiTraPBP} and SGNet~\cite{Wang2021StepwiseGN} to further improve future trajectory prediction performance.  Furthermore, to further evaluate the effectiveness of modeling different modes separately, we design a data-driven validation protocol to compare the predicted future trajectories with real-world data from the dataset spatio-temporally. \rev{Specifically, we further segment the video clips based on the GPS speed of ego vehicle then match all predicted trajectories with nearby ground-truth within the scene,} and we show that our method results in up to a 4.73 \% decrease in frame-wise displacement error over previous work.

We summarize our contributions as follows:
\begin{enumerate}
    \item We propose the Pedestrian Interaction Module (PIM) which leverages pedestrians' behavior, through gesture and gaze,  to learn a behavior-aware interaction representation between pedestrians and their surroundings.
    \item We propose Mode-aware Trajectory Predictor (MTP) to model the multimodal future trajectory distributions separately based on pedestrians' crossing behavior.
    \item We introduce a novel data-driven validation protocol to assess the validity of multimodal trajectory prediction. 
\end{enumerate}

\section{Related work}

\subsection{Pedestrian Trajectory Prediction}
\emph{Bird's-eye view.}
Predominant approaches involve predicting trajectories from bird’s-eye-view videos captured by surveillance cameras or drones. Prior research has focused on multi-agent scenarios, learning diverse trajectories, goal, or group behavior to model the interaction between agents~\cite{Yuan2021AgentFormerAT,Chib2024MSTIPIA,Gu2022StochasticTP,Karim2023DESTINEDG,Xu2022GroupNetMH,Mao2023LeapfrogDM,Wong2023SocialCircleLT}. Another stream of work~\cite{Crosato:TIV2023,Crosato:SVO} is based on modeling vehicle-pedestrian interactions based on Reinforcement Learning and simulation. These approaches typically model interactions based on pedestrians' positions and their relative movement within the observation horizon, using distance-based cues.
However, pedestrians' non-verbal communications with others, such as gestures and nodding, can strongly indicate their intended behaviors and trajectory pattern. Such approaches cannot effectively leverage these visual cues to predict future trajectories due to viewpoint limitations.

\emph{3D view.}
To more accurately model spatiotemporal relationships among road users~\cite{Crosato:VR}, recent research has begun collecting datasets with 3D annotations. MuPoTS-3D~\cite{MuPoTS-3D} and JRDB-GlobMultiPose (JRDB-GMP)~\cite{Jeong:CVPR2024} provide 3D pedestrian poses estimated from video frames, but the resulting motion sequences are often noisy. Wang et al.~\cite{wang:multiPpl} synthesized three-person interactions by combining single- and two-person motion sequences from the high-quality CMU Motion Capture Database~\cite{cmu_mocap}. Although the motion quality is high, the synthesized interactions do not fully reflect real-world pedestrian behavior. More recently, Waymo-3DSkelMo~\cite{Waymo-3DSkelMo} leveraged 3D human shape and motion priors to refine noisy LiDAR data from the Waymo Open Dataset Perception Benchmark~\cite{waymo_open_dataset}, increasing the number of 3D poses from approximately 8,000 to over two million. However, these datasets focus exclusively on pedestrians. Zhu et al.~\cite{Zhu:ICRA2026} addressed this limitation by extending Waymo-3DSkelMo with vehicle information to model vehicle–pedestrian interactions.

\emph{Ego-centric view.}
Ego-centric methods rely on data from an on-board moving camera. Previous methods use multi-stream inputs~\cite{Su2022CrossmodalTB,Saadatnejad2023SocialTransmotionPH,Rasouli2022PedFormerPB,Cadena2022PedestrianG,Rasouli2023ANB,Lian2024HierarchicalTR,Yagi2017FuturePL}, while some works utilize historical trajectories and goal guidance~\cite{Yao2020BiTraPBP,Wang2021StepwiseGN} to achieve notable stochastic results. BR-GAN~\cite{9851641} uses YoloV3 on bird-eye-view video clips to identify pedestrian behavior as walking, grouping and standing then combine it with other temporal features in a late-fusion paradigm. TAGRN-SAR~\cite{Chiu2025TemporalAW} utilizes action intents into autoregressive trajectory decoding. Considering actions reflect the dynamics of pedestrians, the actions combined with Gaussian noise are aggregated with encoder output to enhance the model's awareness of future dynamics.
Compared to aforementioned methods, we not only introduce gaze and gesture, which are highly related to interaction instead of using dynamic-related actions, but also utilize crossing actions as pedestrians' modes, enhancing the model's ability to perceive and model the multimodality of future trajectory.

\emph{Multimodality in trajectory prediction.}
Mainstream methods~\cite{Salzmann2020TrajectronDT,Mao2023LeapfrogDM,Gu2022StochasticTP,Yao2020BiTraPBP,Wang2021StepwiseGN} adopt generative models, such as CVAE or diffusion architecture for stochastic prediction, and incorporate losses for encouraging the model to sample diverse trajectories with a single distribution. 
Although these methods achieve state-of-the-art performance, we consider that a unimodal distribution is insufficient to capture the multimodality of pedestrian motion properly. Inspired by recent advancements in multimodal vehicle trajectory prediction~\cite{Zhang2024DeMoDM,wang2025c2f,Choi2022HierarchicalLS}, we propose to explicitly define and separate multiple mode distributions, enabling the model to better understand the multimodal future of pedestrian. We further validate through extensive experiments that our approach mitigates the above issue.

\section{Methodology}
\begin{figure*}[htb]
  \centering
  \includegraphics[width=\textwidth]{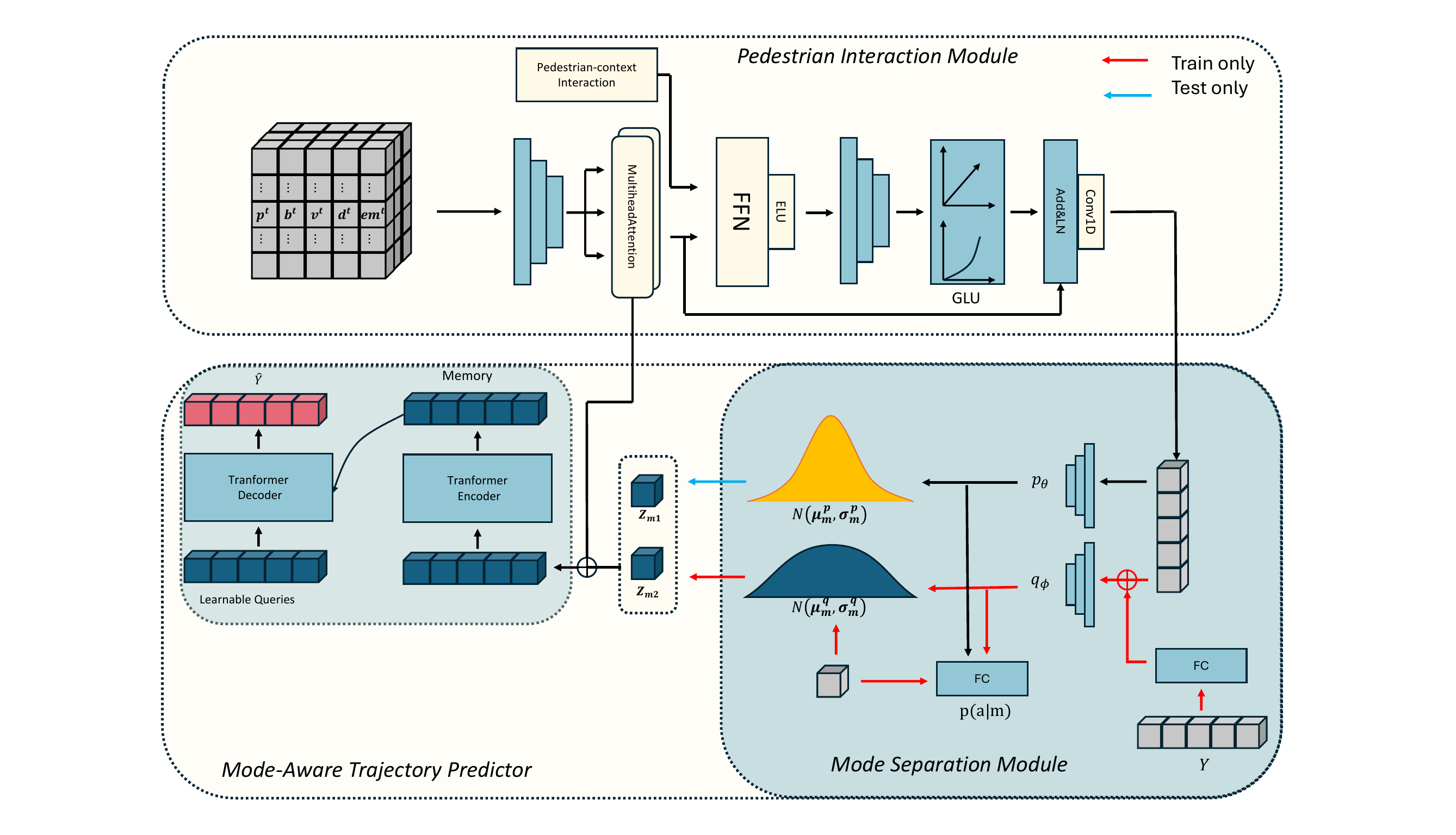}
  \caption{Our proposed framework: Pedestrian Interaction Module, Mode-aware Trajectory Prediction module and $m$ decoders for each separated mode. {\color{red}Red line} denotes training-only while {\color{blue}blue line} denotes test-only.}
  \label{fig:framework}
\end{figure*}

\subsection{Problem Formulation}

\rev{We formulate trajectory prediction as a sequence generation problem which generates the future
trajectory of an agent based on the observed history. For a target agent, given its encoder input sequence
$\mathbf{X}^{t}=\{\mathbf{x}^{t-{\mathrm{T_h}}+1},\ldots,\mathbf{x}^{t-1},\mathbf{x}^{t}\}$
over ${\mathrm{T_h}}$ frames, we aim to predict its future trajectory
$\mathbf{Y}^{t}=\{\mathbf{y}^{t+1},\mathbf{y}^{t+2},\ldots,\mathbf{y}^{t+T_{\mathrm{f}}}\}$
over ${\mathrm{T_f}}$ future frames, where $\mathbf{x}^{t}$ denotes the input feature vector at the $t$-th time step, and $\mathbf{y}^{t}$ denotes the pedestrian bounding-box coordinates at the $t$-th time step.}

To model the interaction between \rev{the target} and the surroundings, multiple modalities are considered in this work:
\begin{enumerate}
    \item pedestrian's normalized bounding-box coordinates $\mathbf{p}^{t}$,
    \item pedestrian's displacement relative to the last observed time step $\mathbf{d}^{t}$,
    \item pedestrian's velocity $\mathbf{v}^{t}$, defined as the displacement of the pedestrian with respect to the previous time step.
\end{enumerate}

The above features are computed on the image plane \rev{$(x_{1}^{t}, y_{1}^{t}, x_{2}^{t}, y_{2}^{t})$, corresponding to top-left and bottom-right corners of the pedestrian bounding boxes.}

In addition, we use behavioral features $\mathbf{b}^{t} $ of pedestrian which have not been explored in the previous work:
\begin{enumerate}
\setcounter{enumi}{3}
    \item \textit{gaze} and \textit{gestures} \rev{$\mathbf{b}^{t} = (g^{t}, \ell^{t})$,}, depending on availability in the dataset.
\end{enumerate}
Furthermore, we include
\begin{enumerate}
\setcounter{enumi}{4}
    \item ego-vehicle motion \rev{  $\mathbf{em}^{t} = (s^{t},\, s_{x}^{t},\, s_{z}^{t})$, 
    i.e., the vehicle GPS speed and its decomposed velocities along the $x$- and $z$-axes, and
    \item semantic context $\mathbf{sc}^{{\mathrm{T_h}}}$ at the last observation,}
   represented by the observed semantic segmentation map classified into four categories: pedestrians, motorcyclists/bicyclists, vehicles, and static context \rev{which dynamically interact with the target in the scene.}
\end{enumerate}

\subsection{Modelling Multimodal Pedestrian Manoeuvres (MMPM)}

Figure~\ref{fig:framework} illustrates our proposed mode-aware framework for pedestrian trajectory prediction, MMPM, where we incorporate new behavioral features into interaction modeling in the Pedestrian Interaction Module (PIM) (Section~\ref{sec:PIM}) and separately model future trajectory distributions under different pedestrian modes in the Mode-aware Trajectory Predictor (MTP) (Section~\ref{sec:MTP}). 


\subsection{Pedestrian Interaction Module (PIM)} \label{sec:PIM}
We design PIM to capture interaction among pedestrian, ego-vehicle and context, by integrating new gesture and gaze features into other motion cues. 
It is suggested in the literature~\cite{Shu2018PerceptionOH,Salzmann2020TrajectronDT} that the interaction relies on low-level motion, such as direction, speed, relative position, etc. However, we argue that interaction is influenced not only by such motion cues but also by behavior that reveal the pedestrian's future intention. 
For instance pedestrian behavior such as nodding or hand gestures demonstrates the communication with other agents and hint the intention. To model the influence of behavior, PIM is an encoder based on attention which consists of two components: a pedestrian-vehicle interaction block, and a pedestrian-context interaction block.

Motivated by~\cite{Rasouli2022PedFormerPB}, the pedestrian-context interaction block aims to aggregate scene information based on the pedestrian historical trajectory and and ego-vehicle motion. We split the resized semantic map into $n$ patches of size $p_z \times p_z$. After applying multi-head attention over the patch tokens, we feed the attended context tokens together with the last hidden state of \rev{ displacement $\mathbf{d}^{\,i}$, and ego-vhicle motion $\mathbf{em^{\,i}}$} into a global attention unit to obtain spatial interaction features for the target pedestrian, namely context vector $C$:
\begin{align}
{E} &= \mathrm{MultiHead}\!\left(S_m\right) + S_m \nonumber\\
{g} &= \mathrm{softmax}\!\left({h} \odot \mathrm{W}_1 {E}\right) \nonumber\\
{C} &= \tanh\!\left(\mathrm{W}_c\left[{g} \oplus {h}\right]\right)
\end{align}
$W$ denotes a Feed Forward Network~(FFN), MultiHead denotes a multi-head attention unit, $ S_m $ represents embedding of the segmentation patches, \rev{$h$ is the final hidden state of a gated recurrent unit~(GRU)}, and $\oplus$ refers to concatenation.

\begin{figure}[htb]
  \centering
  \includegraphics[width=\columnwidth,trim={0 3.5cm 0 3.5cm},clip]{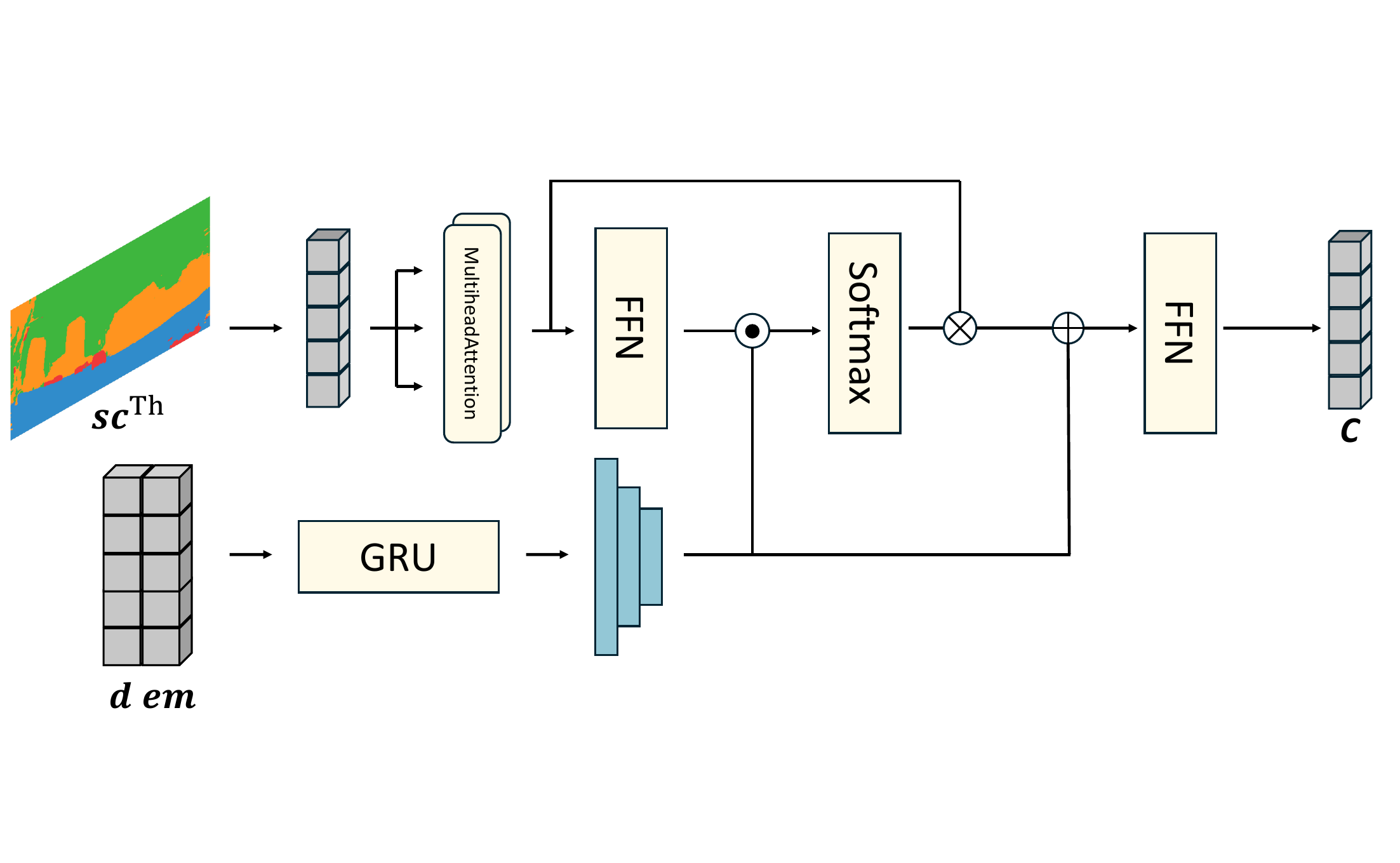}
  \caption{The architecture of Pedestrian-context interaction component.}
  \label{fig:PCI}
\end{figure}

In the pedestrian-context interaction block~(Figure~\ref{fig:PCI}), ego-vehicle and pedestrian motion embeddings are concatenated first then pass through a multi-head attention layer.
The embeddings are further refined using a Gated Residual Network(GRN)~\cite{Lim2019TemporalFT} network to better learn the spatial-temporal dependencies. This is comparable to \cite{Chiu2025TemporalAW}, but we improve it by leveraging $C$ to condition the embeddings. The context vector determines the extent to which information in the embeddings is transformed, allowing the model to control the temporal feature update according to the pedestrian-context interaction, rather than only relying on temporal features for selection. We then employ a global pooling layer, implemented as 1D convolution, to encode the multimodal latent representation: 
\begin{align}
{A} &= \mathrm{MultiHead}\!\left(e\right) + e \nonumber\\
I &= \mathrm{Conv1D}(\mathrm{LayerNorm}\!\big({A}+\mathrm{GLU}(n_1)\big)), \nonumber\\
\eta_1 &= \mathrm{MLP}(\eta_2), \nonumber\\
\eta_2 &= \mathrm{ELU}\!\big(\mathrm{FFN}({A})+\mathrm{FFN}(c)\big).
\end{align}
where $e$ denotes the output of embedding layers,
where MLP is a three-layer Multilayer Perceptron~(MLP), ELU refers to Exponential Linear Unit activation function~\cite{Clevert2015FastAA}, and GLU refers Gated Linear Units~\cite{Shazeer2020GLUVI}.

\subsection{Mode-aware Trajectory Predictor} \label{sec:MTP}
\subsubsection{\rev{Mode separation}} 
\rev{Considering the future trajectories vary, and previous works directly map the history prior and aggregate posterior to the same distribution, which are insufficient to fully model multimodality of trajectory,}
we define the crossing action of pedestrian, i.e, crossing or non-crossing as modes $M$, and propose a Mode-aware Trajectory Predictor that generates future trajectory distributions $p(M \mid \mathbf{X}), {M}=\{m_i \mid i=1,2\}$ with probabilities $p(a\mid m_i)$, where the two modes correspond to \emph{crossing} and \emph{non-crossing} respectively. The overall trajectory distributions are represented as:
\begin{align}
p(\mathbf{Y}\mid \mathbf{X})
&= \sum_{m=1}^{2} p(m_i \mid \mathbf{X})\, p(\mathbf{Y}\mid m_i,\mathbf{X})\, p(a\mid m_i).
\end{align}

We exploit a CVAE framework to learn
multimodal future trajectories. Specifically, we employ a recognition network
$q_{\phi}(M \mid \mathbf{X},\mathbf{Y})$, a prior network
$p_{\theta}(M \mid \mathbf{X})$, where $\phi$ and $\theta$ denote the parameters of these networks.

The ground-truth future trajectory $\mathbf{Y}$ is provided to the recognition network together with $\mathbf{X}$ in the training phase. In details, the embedding $emb_y$ are concatenated with $I$ and fed into an MLP, which outputs the parameters of Gaussian distributions ${N}(\boldsymbol{\mu}^{\,q}_{m},\boldsymbol{\sigma}^{\,q}_{m})$ for each mode $m\in {M}$, while the prior network only takes $e$ as input and outputs ${N}(\boldsymbol{\mu}^{\,p}_{m},\boldsymbol{\sigma}^{\,p}_{m})$.
We sample ${z}_{mi}$ from the recognition distributions and decode future trajectories
conditioned on $\mathbf{X}$.

At inference time, $\mathbf{Y}$ is unavailable. Therefore, the decoder samples ${z}_{mi}$ from each mode's distribution ${N}(\boldsymbol{\mu}^{\,p}_{m},\boldsymbol{\sigma}^{\,p}_{m})$ from prior network $p_{\theta}$, and then decodes the corresponding future trajectories $\mathbf{\hat{Y}^{}}$.

\subsubsection{Decoder}To alleviate error accumulation and ensure the mode consistency during decoding, we employ $M$ query-style decoder~\cite{Carion2020EndtoEndOD} for samples from each mode instead of a shared weight decoder. 
Before entering the Transformer encoder, $\mathbf{X}$ and ${z}_{mi}$ are concatenated as condition, and the decoder then uses a fixed length of learnable queries to attend to the encoder outputs. 

\subsubsection{Auxiliary crossing prediction and probability-guided sampling}
To explicitly differentiate modes, we assume that different modes should reflect whether a pedestrian will cross the road. We therefore introduce an auxiliary action prediction task. We use $q_{\phi}(M \mid \mathbf{X},\mathbf{Y})$ or $p_{\theta}(M \mid \mathbf{X})$ as input to an MLP to predict the crossing probability for weighting the sampling process.

During the training, regardless of the predicted probability, we use the ground-truth action label to select
the correct mode distribution and its decoder for sampling and decoding. This strategy encourages each branch to specialize in a single semantic mode and prevents mode-collapse. At inference time, we sample from all mode distributions according to the predicted crossing probability to ensure explicit multimodal outputs. Given a total of $k$ samples and a predicted crossing probability
$P_c$ 
we draw $ Pc\cdot k $ latent samples from the crossing-mode distribution and $ k - Pc\cdot k$ 
samples from the non-crossing-mode distribution. The
decoded trajectories from both modes are then merged as the final prediction. We show that this sampling yields more valid multimodal samples compared with existing methods in Section~\ref{sec:validty}.

\subsection{Objectives}
As discussed above, we predict future displacements
relative to the position $\mathbf{d}^{\,t}$, instead of directly predicting future absolute positions or frame-wise velocity. This helps reduce initial misalignment
, and also mitigates overall trajectory drift. We apply the Best-of-Many RMSE loss\cite{Bhattacharyya2018AccurateAD} as the reconstruction term ${L}_{\mathrm{rec}}$ to encourage diversity, and a KL divergence term ${L}_{\mathrm{kl}}$ to bridge the gap between the prior network $p_{\theta}(M \mid \mathbf{X})$ and the recognition network $q_{\phi}(M \mid \mathbf{X},\mathbf{Y})$ which is a closed form solution. The cross-entropy loss ${L}_{\mathrm{ce}}$ is used for the action prediction head. The KL divergence loss and the overall loss $L$ are formulated as follow:
\begin{align}
{L}_{\mathrm{KL}}
&= \sum_{m \in {M}} \omega_m \,
\mathrm{KL}\!\left(q_{\phi}\!\left(z \mid X,Y,m\right)\,\big\|\,p_{\theta}\!\left(z \mid X,m\right)\right)
\end{align}

\begin{align}
{L}= \alpha\,{L}_{\mathrm{rec}}+ \beta\,{L}_{\mathrm{ce}}+\gamma\,{L}_{\mathrm{kl}}
\end{align}
where $\alpha,\beta,\gamma$ are the weights for each loss term. 

\section{Experiments}

\subsection{Datasets and Metrics}
We evaluate our approach on two public ego-centric datasets, PIE~\cite{Rasouli2019PIEAL} and JAAD~\cite{Kotseruba2016JointAI}, following the default train/test splits. Since ego-motion is not available in JAAD, we use the driver's action annotation as an alternative. We sample the sequences at 30 FPS with a window overlap ratio of $0.5$, using $0.5$s of observation to
predict $1.5$s into the future. Following~\cite{Yao2020BiTraPBP}, we evaluate our method on the image plane in pixels using the
following metrics: MSE on bounding-box coordinates, reported at $0.5$s, $1.0$s, and $1.5$s to reflect short-term to
long-term prediction performance. We also report $C_{MSE}$ and $CF_{MSE}$ at $1.5$s, which measure the average displacement and the final displacement of
the bounding-box center coordinates, respectively. 

\subsection{Implementation Details}

We set all embedding layers to $d=64$, and mixed latent
variable size to $d_{{m}}=64$, compressed by a $3$-layer MLP.  We use 4 attention heads for all multi-head attention and Transformer blocks, and the depth of Transformer is set to 2. The loss weights $\alpha,\beta,\gamma$ are set to 1, 0.1, 0.1, respectively, where ${L}_{\mathrm{kl}}$ is activated after a warm-up of the first $10\%$ of training. We train the model with batch size $B=32$, terminated after $100$ epochs, using a learning rate of $\eta=5\times 10^{-4}$ with a $10\%$ cosine warm-up schedule and the AdamW~\cite{Loshchilov2017FixingWD} optimizer.

\subsection{Quantitative Results}
We conducted extensive experiments to compare our proposed model with previous state-of-the-art pedestrian trajectory prediction
methods on ego-centric datasets, including Bitrap~\cite{Yao2020BiTraPBP}, SGnet~\cite{Wang2021StepwiseGN}, ABC+~\cite{Halawa2022ActionbasedCL}, ENCORE~\cite{Rasouli2023ANB}, and TAGRN-SAR~\cite{Chiu2025TemporalAW}, and results are reported in Table~\ref{tab:pie}. 
Bottom rows show the results of our best model and variants for ablation. Our-NM and Our-NB refer to our model \textit{without} MTP and \textit{without} incorporating behavior features. Compared with previous methods, our model without MTP outperforms the state-of-the-art TAGRN-SAR, achieving a $4\%$ improvement in
$CF_{MSE}$, and further reduces the error by $4.17\%$ after introducing MTP, with more performance gains on the other metrics. In particular, the 1.5s prediction error $C_{MSE}$ is improved by $25\%$ and $7.41\%$, respectively. The incorporation of behavior features also significantly improves our method, leading to $25\%$  and $14.01\%$ gains in short-term prediction and final displacement compared to our model without behavior features.
On the JAAD dataset, Table~\ref{tab:jaad} reports that our model consistently achieves state-of-the-art performance, obtaining up to a $35\%$ improvement in short-term prediction without the module. Both $CF_{MSE}$ and
$C_{MSE}$ are further reduced, 
outperforming the top-performer ABC+. Overall, our framework demonstrates superior
short-term prediction performance compared with previous methods, while the MTP improves all metrics and is particularly beneficial for long-term prediction on JAAD, confirmed by more accurate predictions at $1.5$s.

To further assess the 
effectiveness of modeling each mode separately in our proposed method, we plug our MTP module into both Bitrap and SGnet, without modifying any other layers or replacing the decoder. As in Table~\ref{tab:mode}, we observe consistent improvements across all metrics for both baselines, although the performance gains are slightly smaller than those achieved by our model 
after introducing MTP. We assume that this is because the relatively simple encoders in Bitrap and SGnet cannot capture pedestrian interactions as effectively as ours, limiting the benefit of introducing MTP. Nevertheless, these results indicate that our MTP can be a model-agnostic component to improve existing methods, and explicitly modeling pedestrians' modes helps models predict trajectories more accurately.
\begin{figure*}[!t]
  \centering
  \includegraphics[width=\textwidth]{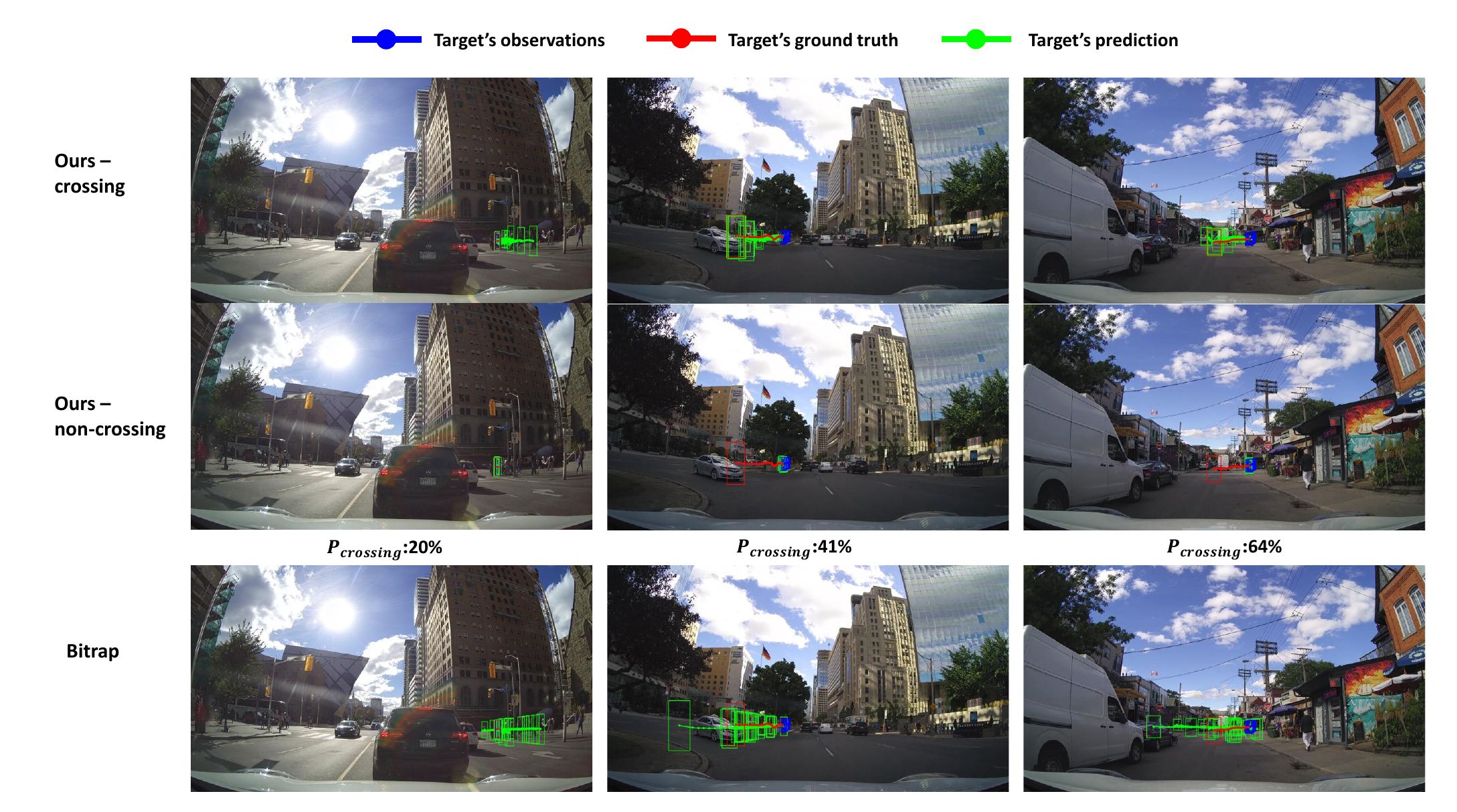}

  \includegraphics[width=\textwidth]{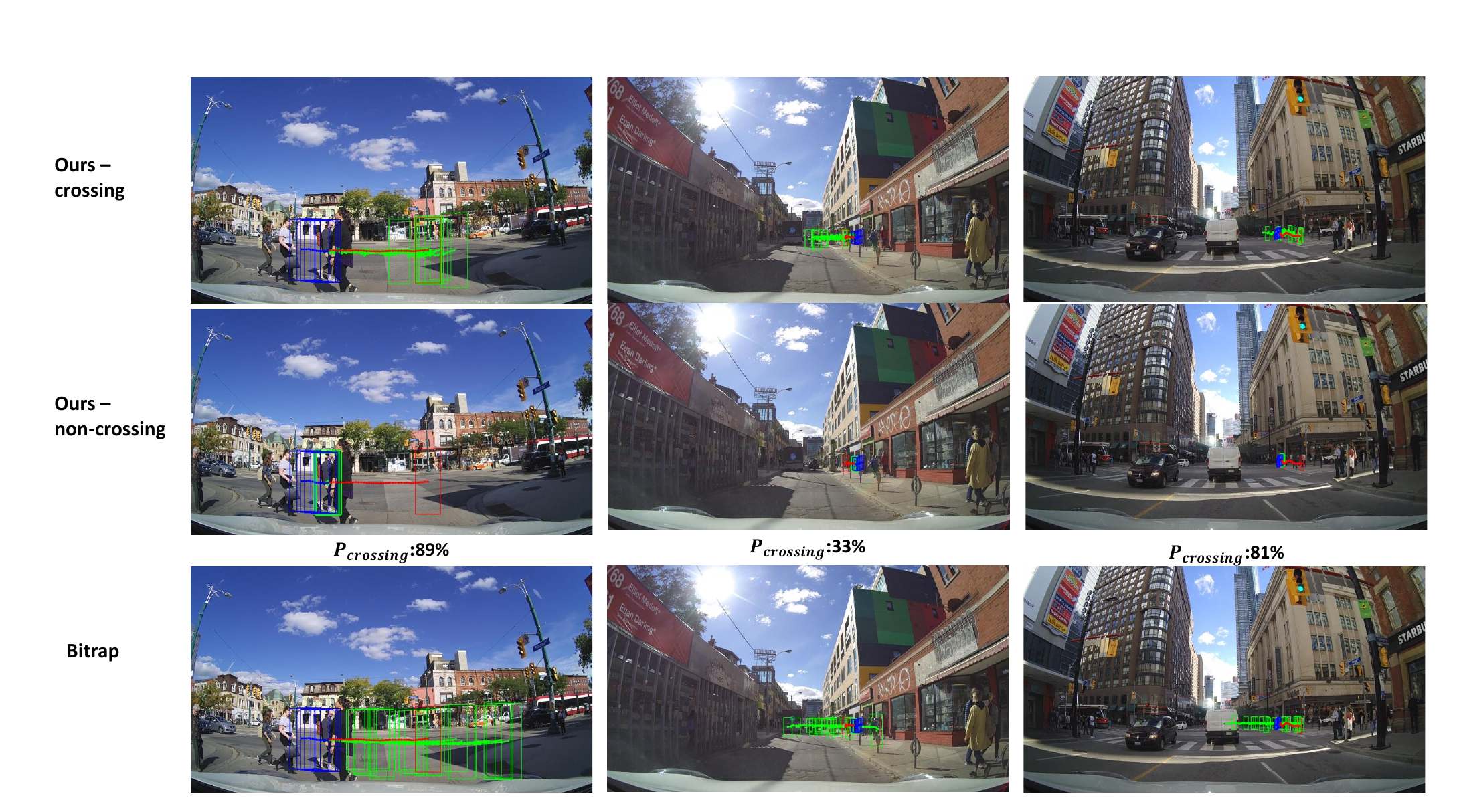}
  \caption{Qualitative results from 2 different scenarios (Top and bottom). For each scenario, the first row demonstrates the prediction of crossing mode in our method, and the second row shows non-crossing prediction. All sampled trajectories of Bitrap are plotted together in the third row. The colors correspond to pedestrian's {\color{blue} historical trajectory}, {\color{red}future ground truth}, {\color{green} prediction}.} 
  \label{fig:qualitative}
\end{figure*}
\begin{table}[!t]
\centering
\caption{Comparison of stochastic results (20 samples) to SOTA on the PIE dataset.Lower is better for all metrics and the best results are \textbf{bold}.}
\label{tab:pie}
\setlength{\tabcolsep}{8pt}
\begin{tabular}{lccccc}
\hline
 & \multicolumn{3}{c}{$MSE$} & $C_{MSE}$ & $CF_{MSE}$ \\
\hline
 & \textbf{0.5s} & \textbf{1s} & \textbf{1.5s} & \textbf{1.5s} & \textbf{1.5s} \\
\hline
BiTrap-NP~\cite{Yao2020BiTraPBP}    & 23 & 48 & 102 & 81 & 261 \\
SGNet-ED~\cite{Wang2021StepwiseGN}     & 16 & 39 &  88 & 66 & 206 \\
ABC+~\cite{Halawa2022ActionbasedCL}         & 16 & 38 &  87 & 65 & 191 \\
ENCORE~\cite{Rasouli2023ANB}       & 15 & 33 &  70 & 49 & 155 \\
TAGRN-SAR~\cite{Chiu2025TemporalAW}    & 16 & 31 &  61 & 36 & 100 \\
\hline\hline
Ours-NB      & 16 & 34 &  63 & 36 &  107 \\
Ours-NM      & 13 & 31 &  56 & 27 &  96 \\
Ours         & \textbf{12} & \textbf{29} &  \textbf{51} & \textbf{25} &  \textbf{92} \\
\hline
\end{tabular}
\end{table}

\begin{table}[!t]
\centering
\caption{Comparison of stochastic results (20 samples) to SOTA on the JAAD dataset.}
\label{tab:jaad}
\setlength{\tabcolsep}{8pt}
\begin{tabular}{lccccc}
\hline
 & \multicolumn{3}{c}{${MSE}$} & $C_{MSE}$ & $CF_{MSE}$ \\
 \hline
 & \textbf{0.5s} & \textbf{1s} & \textbf{1.5s} & \textbf{1.5s} & \textbf{1.5s} \\
\hline
BiTrap-NP~\cite{Yao2020BiTraPBP}   & 38 & 94 & 222 & 177 & 565 \\
SGNet-ED~\cite{Wang2021StepwiseGN}    & 37 & 86 & 197 & 146 & 443 \\
ABC+~\cite{Halawa2022ActionbasedCL}         & 40 & 89 & 189 & 145 & 409 \\
ENCORE~\cite{Rasouli2023ANB}       & 32 & 85 & 210 & 167 & 554 \\
\hline\hline
Ours-NM      & 26 & 81 & 197 & 145 & 448 \\
Ours         &  \textbf{24} &  \textbf{73} &  \textbf{177} & \textbf{141} &  \textbf{401} \\
\hline
\end{tabular}
\end{table}

\begin{table}[!t]
\centering
\caption{Results of introducing Mode Separation Module to SOTA.}
\label{tab:mode}
\setlength{\tabcolsep}{8pt}
\begin{tabular}{lccccc}
\hline
 & \multicolumn{3}{c}{$MSE$} & $C_{MSE}$ & $CF_{MSE}$ \\
\hline
 & \textbf{0.5s} & \textbf{1s} & \textbf{1.5s} & \textbf{1.5s} & \textbf{1.5s} \\
\hline
BiTrap-NP~\cite{Yao2020BiTraPBP}    & 23 & 48 & 102 & 81 & 261 \\
SGNet-ED~\cite{Wang2021StepwiseGN}     & 16 & 39 &  \textbf{88} & 66 & 206 \\

\hline\hline
BiTrap-NP-M  & 17 & 46 &  99 & 77 & 252 \\
SGNet-ED-M   & \textbf{13} & \textbf{34} &  91 & \textbf{63} & \textbf{196} \\
\hline
\end{tabular}
\end{table}

\begin{table}[!t]
\centering
\caption{Per-frame error validity evaluation to baselines on PIE dataset. }
\label{tab:valid}
\setlength{\tabcolsep}{8pt}
\resizebox{\columnwidth}{!}{
\begin{tabular}{lccc}
\hline
 & \textbf{All Scenes MSE} & \textbf{Two Side MSE} & \textbf{One Side MSE} \\

\hline
BiTrap-NP~\cite{Yao2020BiTraPBP} & 37.39 & 35.03 & 39.74 \\
SGNet-ED~\cite{Wang2021StepwiseGN}  & 36.75 & 35.13 & 38.39 \\
\hline
Ours   & \textbf{35.62} & \textbf{33.67} & \textbf{37.57} \\
\hline
\end{tabular}
}
\end{table}
\subsection{Qualitative Results}
Fig.~\ref{fig:qualitative} shows a qualitative comparasion on PIE between our best model and Bitrap whose source is available. To investigate the performance on each mode, we force the model to predict 10 crossing and 10 non-crossing trajectories among 20 generated samples while showing the actual crossing probability which indicates the sampling ratio. The results show that the modes we defined consistently generate clearly different trajectories, corresponding to being stationary or crossing the road. Besides, within each mode, the final destinations are more concentrated or clustered, indicating
that the learned mode effectively constrains the sampling process. In contrast, although Bitrap can produce diverse trajectories, some samples appear to lie between two sampled trajectories which are not valid in the scene. This may be caused by the limitations of sampling from mixed modes. The results demonstrate our method's capacity to separate and model pedestrians' motion modes and pedestrians' intention awareness, resulting in more valid samples.

\subsection{Data-driven Trajectory Validation}
\label{sec:validty}
To verify whether modelling pedestrian modes separately can generate more realistic trajectories, we propose a data-driven trajectory realism evaluation protocol. Instead of only comparing predictions against the paired future trajectory of the target pedestrian, we measure the distance between each prediction and spatio-temporally consistent real-world trajectories observed in the same scene. Since valid reference trajectories should correspond to similar scene contexts and traffic conditions, we enforce strict spatio-temporal consistency when constructing trajectory candidates. Specifically, we segment the test videos into clips of different durations according to the ego-vehicle GPS speed, while preserving the original dataset split. 
We empirically use $10$~km/h as the speed threshold~\cite{Rasouli2023ANB} to divide long videos into clips. Predictions for each pedestrian within a clip are then matched against the available reference trajectories in that clip.

Specifically, taking the last observed position as the center, we search all ground-truth future trajectories within a radius of $150$ pixels, and compare them against all predicted trajectories by the minimum displacement error. To be consistent with the benchmark, we also adopt $\mathrm{MSE}$ as the metric. Considering its sensitivity to outliers, we remove errors outside the inter-quartile range from all best matches. We further categorize each circle as two-side and one-side , depending on if both crossing and non-crossing ground-truth trajectories are available and matched against predictions. As shown in Table~\ref{tab:valid}, our method achieves smaller errors against real samples than the baselines, and remains closer to ground-truth trajectories in two-side circles. The scarcity of comparable ground-truth trajectories within the search radius and the Best-of-many loss term, may cause this higher error compared with the prediction task, as the rest of other samples are not regularized while there are only a few ground truth to be matched.

\section{Conclusion}
\rev{In this work, we proposed MMPM, a mode-aware trajectory prediction framework based on a novel mode-aware modeling that handles the multimodality of pedestrian future trajectories in autonomous driving. Before mapping prior and posterior, MMPM separates different behavior modes in the future distribution and samples corresponding trajectories. Experimental results on PIE and JAAD datasets show that our method achieves state-of-the-art performance over previous trajectory prediction methods, and qualitative results demonstrate the effectiveness of explicit mode modeling. We also introduce a new experimental protocol to validate the validity of stochastic trajectory prediction. The results suggest our method is able to generate more valid trajectories, which mitigates the issue of distribution misalignment.}

There are limitations in our current approach that suggest potential directions for future work. Our method \rev{still maps future and history to a normal Gaussian after separating the modes, and the modes were defined based on} the crossing annotations provided in PIE and JAAD. Since entire long pedestrian sequences are labeled by such annotations, the crossing behavior may not have started or may have already ended within the time horizon. The data split of our and previous methods does not consider this mislabeling when modeling action and trajectory prediction jointly. In addition, binary modes may not fully cover the multimodality of pedestrian motion yet. Future work could explore alternative approaches for identifying modalities, adopting either predefined or dynamically separated modalities to reduce reliance on annotations while introducing more diverse modes \rev{, or improve the trajectory prior representation to better model such one-to-many mapping problem.}


\bibliographystyle{IEEEtran}
\bibliography{IEEEabrv,ref}

\end{document}